%% file: bayes.tex
\documentclass[letterpaper, 10 pt, conference]{ieeeconf}
\IEEEoverridecommandlockouts                              

\usepackage{times}
\usepackage{graphics} 
\usepackage{epsfig} 
\usepackage{amsmath} 
\usepackage{amssymb}  

\usepackage{amsmath,bm,times}
\usepackage[usenames,dvipsnames,table]{xcolor}
\usepackage{array}
\usepackage{colortbl}	
\usepackage{url}
\usepackage{mathtools}
\usepackage{tikz}
\usetikzlibrary{matrix,decorations.pathreplacing}

\usepackage{multicol}

\usepackage{flushend}

\newtheorem{property}{Property}

\newtheorem{problem}{Problem}

\newtheorem{definition}{Definition}

\newtheorem{remark}{Remark}

\title{ \LARGE \bf Inverse, forward and other dynamic computations
computationally \\ optimized with sparse matrix factorizations}

\author{Francesco Nori$^{1}$ 
\thanks{$^{1}$Francesco Nori is with iCub Facility,
        Istituto Italiano di Tecnologia, Italy.
        {\tt\small francesco.nori@iit.it}. This paper was supported by the FP7 EU projects CoDyCo (No. 600716 ICT 2011.2.1 Cognitive Systems and Robotics), and An.Dy funded by the European Union's Horizon 2020 Research and Innovation Programme under Grant Agreement No. 731540.}
        }%

\begin{document}

\maketitle
\thispagestyle{empty}
\pagestyle{empty}

\begin{abstract}
We propose an algorithm to compute the dynamics of 
articulated rigid-bodies with different sensor distributions.
Prior to the on-line computations, the proposed
algorithm performs an off-line optimisation step to simplify the 
computational complexity of the underlying solution. 
This optimisation step consists in formulating the dynamic
computations as a system of linear equations. The computational 
complexity of computing the associated solution is reduced 
by performing a permuted $LU$-factorisation with off-line optimised permutations.
We apply 
our algorithm  to solve classical 
dynamic problems: inverse and forward dynamics.
The computational 
complexity of the proposed solution
is compared to `gold standard' algorithms: recursive Newton-Euler and articulated 
body algorithm. It is shown that our
algorithm reduces the number of floating point operations with respect
to previous approaches. We also evaluate the numerical complexity of our algorithm 
by performing tests on dynamic computations for which no gold standard is available. 

\end{abstract}

   
\input{2_method.tex}


\bibliographystyle{IEEEtran}
\bibliography{bayesbiblio}

\end{document}

%% file: 2_method.tex

\section{Previous works}

Real-time control of complex robots (such as humanoids) asks for efficient ways of computing position, velocity, acceleration and applied wrenches on all the bodies composing the robot articulated chain. Within this paper, we will
refer to these quantities with the term \emph{dynamic variables} even though they include positions and velocities
which are strictly speaking kinematic. Efficiently computing the dynamic variables of a (possibly free-floating) robot
is nowadays a solved theoretical problem. Several state of the art algorithms \cite{Featherstone2008} have a
computational complexity which scales linearly in the number of rigid links composing the robot. All these efficient
algorithms can be obtained by exploiting the problem sparsity which derives from the iterative propagation of forces,
torques and accelerations across articulated structures. Classical problems have been extensively optimized from the
computational complexity point of view. As discussed in \cite[Table 10.1]{Featherstone2008} the recursive
Newton-Euler algorithm as implemented in \cite[Algorithm 5.7]{balafoutis1991dynamic} is the most computationally
efficient solution of the inverse dynamics problem. Similarly, the articulated body algorithm as implemented in
\cite[method 3]{Orin1982} is the most computationally efficient solution of the forward dynamics problem. Yet another
relevant problem is the hybrid dynamics solution, which can be efficiently solved with the articulated-body hybrid
dynamics \cite[Section 9.2]{Featherstone2008}.

In this paper we consider the computationally efficient solution of dynamic problems which do not fall within the scope of the inverse, forward and hybrid dynamics even though they are of practical interest. The paper is organized as follows. Section \ref{sec:notation} present the notation. Section \ref{sec:method} presents a specific formulation of the Newton-Euler equations and Section \ref{sec:dyn_constr} discuss the matrix form of these constraints. Section \ref{sec:inverseDynamics} presents the inverse dynamics problem and its solution with the recursive Newton-Euler algorithm (Section \ref{sec:RNE}). Section \ref{sec:fwdDyn} presents the forward dynamics problem and its solution with the articulated body algorithm (Section \ref{sec:ABA}). An alternative but computationally comparable solution of the inverse and forward dynamics is proposed in Section \ref{sec:luRNE} and Section \ref{sec:luAB} respectively. Finally, Section \ref{sec:LUgeneric} extends this solution to dynamic problems which do not fall within the scope of the inverse, forward and hybrid dynamics.

\section{Notation} \label{sec:notation}

Let $\mathcal A$ and $\mathcal B$ be two arbitrary sets with cardinality $|\mathcal A| = m$ 
and $|\mathcal B| = n$. For any element in $a \in \mathcal A$ let's associate a vector $d_a$. 
Similarly, for any pair of elements $a,b$ with $a \in \mathcal A$ and $b \in \mathcal B$, 
let's associate a unique matrix $D_{a,b}$. Let $p = (a_1, \dots, a_m) \in S_{\mathcal A}$ and $q = (b_1, \dots, 
b_n) \in S_{\mathcal B}$ be two permutations of the elements in $\mathcal A$ and $\mathcal B$ 
respectively\footnote{Equivalently, let $p \in S_{\mathcal A}$ and $q \in S_{\mathcal B}$, i.e. 
let $p$ and $q$ be two elements of the symmetric group on $\mathcal A$ and 
$\mathcal B$ respectively.}. We define the permuted vector $\bm d_p$ and the 
permuted matrix $\bm D_{p,q}$ induced by $p$ and $q$ as follows:
\begin{equation}
\bm d_{p} = \begin{bmatrix}  
d_{a_1}\\
\vdots \\
d_{a_m}
\end{bmatrix}, \qquad
\bm D_{p,q} = \begin{bmatrix}  
D_{a_1, b_1} & \dots & D_{a_1, b_n} \\
\vdots & \ddots & \vdots \\
D_{a_m, b_1} & \dots & D_{a_m, b_n}
\end{bmatrix}.
\end{equation}

As for dynamic quantities, in this paper we follow the same notation used in \cite{Featherstone2008}
but we avoid using the bold symbols for matrices, since the bold symbols are reserved for distinguishing 
$D_{a,b}$, $a \in \mathcal A$, $b \in \mathcal B$ from $\bm D_{p,q}$, $ p \in S_{\mathcal A}$,
$ q \in S_{\mathcal B}$; ${v}$ denotes the spatial velocity (a six dimensional vector including angular velocities in the first three components and the rest as linear velocities), ${a}$ denotes the spatial acceleration (again angular and then linear), ${f}$ denotes the spatial force (couples in the first three components and forces in the remaining), $^B {X}_A$ is the motion vector transformation from $A$ to $B$ coordinates and $^B {X}^*_A$ is the analogous transformation for a force vector. Given a vector $v$ and its coordinates $^A v$ in $A$ we denote with $^A \dot {v}$ its temporal derivative in the $A$ coordinates. The Euclidean cross operator $\times$ (on $\mathbb R^3$) is defined as the usual vector product, while for a spatial velocity $v$ the cross spatial cross operator and its dual $\times^*$ are defined as follows:
\begin{eqnarray*}
v \times =   \begin{bmatrix} \omega \\ {\dot p}\end{bmatrix} \times = \begin{bmatrix} \omega \times  & 0 \\ {\dot p} \times  & \omega \times \end{bmatrix}, 
v \times^* = \begin{bmatrix} \omega \\ {\dot p}\end{bmatrix} \times^* =\begin{bmatrix} \omega \times &  {\dot p} \times\\ 0 & \omega \times \end{bmatrix}
\end{eqnarray*}
In the following sections, we restrict the analysis to robots described by kinematic trees (i.e. robots with no kinematic loops) composed of $N_B$ rigid links numbered from $0$ to $N_B$, $0$ being the selected fixed base (world reference frame) and $1$ being the selected floating base (reference rigid body in the articulated chain). Links numbering is obtained by defining a suitable spanning tree where each rigid link is associated to a unique node in the tree. Numbers can be always selected in a topological order so that each node $i$ has a higher number than its unique parent $\lambda_i$ and a smaller number than all the nodes in the set of its children $\mu_i$. Links $i$ and $\lambda_i$ are coupled with the joint $i$ whose joint motion constraints are modelled with $S_i \in \mathbb R^{6 \times 1}$ (therefore without loss of generality we are assuming that all joints have a single degree of freedom). For each rigid link $i$, the system is modelled also by supplying the spatial inertia tensor\footnote{The spatial inertia tensor of the rigid link $i$ has the following form in the link reference frame: $$ I_i = \begin{bmatrix} I_{C,i} + m_i c_i \times c_i^\top & m_i c_i \times \\ m_i c_i \times^\top & m_i I_{3 \times 3} \end{bmatrix},$$ where $I_{C,i}$ is the spatial inertia tensor with respect to the body's centre of mass, $m_i$ is the total mass, $c_i$ is the relative displacement between the centre of mass and the origin of the link reference frame.} $I_i$ and the motion-vector transform from the reference frame of the rigid link $i$ to the reference frame of the rigid link $j$, denoted $^j X_i$. For each link $i$ the considered kinematic variables are: 
\begin{itemize}
\item[] $v_i$: the link spatial velocity,
\item[] $q_i$: the joint $i$ position, 
\item[] $\dot {q}_i$: the joint $i$ velocity,
\end{itemize}
Similarly, the dynamic variables associated to the link $i$ are: 
\begin{itemize}
\item[] $a_i$: the link spatial accelerations, 
\item[] $\ddot {q}_i$: the joint $i$ acceleration, 
\item[] ${\tau}_i$: the joint $i$ torque, 
\item[] $f_i$: the spatial force transmitted to body $i$ from $\lambda_i$,
\item[] $f_i^x$: external forces acting on body $i$.
\end{itemize}
\begin{remark}
All these variables are expressed in body $i$ coordinates including $f_i^x$ which is more often expressed in absolute (i.e. body $0$) coordinates. 
\end{remark}
 

\section{Dynamic constraints} \label{sec:method}

Let's assume that all kinematic quantities, i.e. those depending on $q$ and $\dot {q}$ have been precomputed. In practice, these quantities are the transformation matrices $^j X_i$, $^j X^*_i$ and the 
linear/angular velocities $v_{Ji}$, $v_{i}$. The latter can be efficiently computed with 
the following recursion, propagated from $i = 1$ to $N_B$:

\begin{equation} \label{eq:NEkin}
v_i = \prescript{i}{} {X_{\lambda_i}} {v}_{\lambda_i} + S_i \dot {q_i}, 
\end{equation}

\noindent
The dynamics quantities instead ($\tau$, $\ddot {q}$, $f_1^x, \dots, f_{N_B}^x$, $a_1, \dots, a_{N_B}$, $f_1, \dots,
 f_{N_B}$) have to satisfy the Newton-Euler equations:

\begin{subequations} \label{eq:NEdyn} 
\begin{align} 
a_i    &= \prescript{i}{}{X_{\lambda_i}} a_{\lambda_i} + S_i \ddot {q}_i + c_i, \label{eq:ai} \tag{\theequation $a_i$} \\
\tau_i  & =  S^\top_i f_i \label{eq:taui} \tag{\theequation $\tau_i$}, \\
f_i & =  I_i a_i + \nu_i  -  f_i^x  + \sum_{j \in \mu_i} \prescript{i}{}{X_{j}^*} f_j \label{eq:fi} \tag{\theequation $f_i$}.
\end{align}
\end{subequations}
where we defined $\nu_i = v_i \times^* I_i v_i$ and $c_i = 
v_i \times  S_i \dot {q}_i$.

\subsection{Dynamic constraints in matrix form} \label{sec:dyn_constr}

Let's define 
the set of dynamic variables $\mathcal D = \{ a_{i}$,
$f_{i}$, $\tau_{i}$, $f_{i}^x$, $\ddot {q_{i}}$ $\}_{i=1}^{N_B}$.
We have $|\mathcal D| = 5N_B$. 
The elements in $\mathcal D$ can be thought as generic quantities,
not necessarily expressed in a specific reference frame. As described in Section
\ref{sec:notation}, for any $a \in \mathcal D$, let's associate a vector $d_a$ (e.g. $d_{a_{i}}$ might correspond to the coordinates of $a_{i}$ expressed in a specific 
reference frame). Given a permutation $q \in \mathcal S_{\mathcal D}$, $d_q$ contains an ordered sequence of the elements in $\mathcal D$ expressed in specific reference frame(s).

Equations \eqref{eq:NEdyn} can be seen as a set of equations
which the \emph{vector of dynamic variables} $d_p$ have to satisfy. Let's put
these constraints in matrix form by defining the set of constraints $\mathcal C
 = \{ c_\eqref{eq:ai}$, $c_\eqref{eq:fi}$, $c_\eqref{eq:taui} 
 \}_{i=1}^{N_B}$. We have $|\mathcal C| = 3 N_B$. Given $c \in \mathcal C$ let's define vectors $b_c$ as follows:

\begin{eqnarray*}
 b_{c_\eqref{eq:ai}}
  & = & \left\{ \begin{array}{ll}
  \prescript{i}{}{X_{0}} 
 a_0 + c_i & \mbox{if } \lambda_i=0 \\
 \quad v_i \times S_i \dot {q_i} & \mbox{if } \lambda_i \neq 0
 \end{array} \right., \\
  b_{c_\eqref{eq:fi}} &= & \nu_i,
\end{eqnarray*}
and $b_{c} = 0$ otherwise. Given $c \in \mathcal C$ and $d \in \mathcal D$, let's define
matrices $D_{c,d}$ as follows (where $j \in \mu_i$):

\begin{subequations} \label{eq:Dblocks} 
\begin{align}
  D_{c_\eqref{eq:ai}, a_{i}} & = -1_6, &
  D_{c_\eqref{eq:ai}, a_{\lambda_i}} & = \prescript{i}{}{X_{\lambda_i}}, & \\
  D_{c_\eqref{eq:ai}, \ddot {q_{i}}} & = S_i, &
 & \\[8pt]
  D_{c_\eqref{eq:fi}, f_{i}} & = -1_6, &
  D_{c_\eqref{eq:fi}, a_{i}} & = I_i, \\
  D_{c_\eqref{eq:fi}, f^x_{i}} & = -1_6, &
  D_{c_\eqref{eq:fi}, f_{j}} & = \prescript{i}{}{X_{j}^*}, \\[8pt]
  D_{c_\eqref{eq:taui}, \tau_{i}} & = -1_{n_i}, & 
  D_{c_\eqref{eq:taui}, f_{i}} & = S_i^\top
\end{align}
\end{subequations}
and $ D_{c,d} = 0$ otherwise.

\noindent Given two permutations $p$ and $q$ of the elements in $\mathcal C$ and $\mathcal D$, the permuted matrix $\bm D_{p,q}$ and permuted vectors $\bm d_{q}$, $\bm b_{p}$ satisfy the following linear equation: 
\begin{equation} \label{eq:matRNEA2}
 \bm D_{p,q}(q, \dot {q}) \bm d_{q} + \bm b_{p} (q,{\dot q}) = 0, 
\end{equation}
where we explicitly indicated the dependency on $q$ and $\dot {q}$.

\section{Inverse dynamics} \label{sec:inverseDynamics}

The inverse dynamics problem consists in finding ${\tau_1}$, $\dots$, ${\tau_{N_B}}$ 
which satisfies \eqref{eq:NEdyn} given $\ddot {q}_1$, $\dots$, $\ddot {q}_{N_B}$,  
$f_1^x$, $\dots$, $f_{N_B}^x$. In \cite{Featherstone2008} the problem is formulated
as the computation of the following function:
\begin{eqnarray}
\tau = \mbox{InvD}(\textcolor{gray}{model, q, \dot {q},} \ddot {q}, f_1^x, \dots, f_{N_B}^x).
\end{eqnarray}
In the above equations we grayed out some variables that will not play a role in the following sections,
and can be assumed either to be contestant ($model$) or measured ($q$, $\dot {q}$).

\subsection{Inverse dynamics solved with RNEA} \label{sec:RNE}

An efficient solution of inverse dynamics is given by the recursive Newton-Euler algorithm (RNEA), described hereafter. Equations \eqref{eq:NEkin} and \eqref{eq:ai} are propagated from $1$ to $N_B$ with initial conditions $v_0 = 0$ and $a_0 = -a_g$ which corresponds to the gravitational spatial acceleration vector expressed in the body frame $0$ (null in its first three components and equal to the gravitational acceleration in the last three). Equations \eqref{eq:fi} and \eqref{eq:taui} are propagated from $N_B$ to $1$.

\subsection{Inverse dynamics solved with matrix inversion}

In this Section, we propose a way to solve the inverse dynamics problem with a matrix inversion. Remarkably, \eqref{eq:matRNEA2} represents the set of linear constraints in $ \bm d_q$. Additionally, certain components of $d_p$ are known since $\ddot {q}_1$, $\dots$, $\ddot {q}_{N_B}$,  $f_1^x$, $\dots$, $f_{N_B}^x$ are given: 
\begin{subequations} \label{eq:meas} 
\begin{align} 
\ddot {q}_i & = y_{\ddot {q}_i}, \label{eq:yd2qi} \tag{\theequation $\ddot q_i$} 
\\
f_1^x & = y_{f_i^x}. \label{eq:ydfxi} \tag{\theequation $f^x_i$}
\end{align}
\end{subequations}

These constraints, will extend the set of constraints $\mathcal C$. In particular, we
should define $\mathcal C_{id} = \mathcal C \cup \{ c_\eqref{eq:yd2qi}, c_\eqref{eq:ydfxi} \}_{i=1}^{N_B}$
so that $|\mathcal C_{id}| = 5N_B$. To extend the definition of $D_{c,d}$ and $b_c$ we define:
\begin{subequations} \label{eq:y_RNE}
\begin{align}
  D_{c_\eqref{eq:yd2qi}, \ddot q_{i}} & = 1_{n_i}, & 
  b_{c_\eqref{eq:yd2qi}} & = -y_{\ddot {q}_i},  \\[8pt]
  D_{c_\eqref{eq:ydfxi}, f^x_{i}} & = 1_6, &
  b_{c_\eqref{eq:ydfxi}} & = -y_{f_i^x}, & 
\end{align}
\end{subequations}
and $ D_{c,d} = 0$, $ b_{c} = 0$ otherwise. Again, given two permutations $p_{id}$ and
$q$ of the elements in $\mathcal C_{id}$ and $\mathcal D$, the permuted matrix $ \bm D_{p_{id},q}$ 
and permuted vectors $\bm d_{q}$, $\bm b_{p_{id}}$ satisfy the following linear equation: 
\begin{equation} \label{eq:matRNEA3}
 \bm D_{p_{id},q}(q) \bm d_{q} + \bm b_{p_{id}} (q,{\dot q}) = 0, 
\end{equation}
which is representation of the inverse dynamics problem. In particular, inverting 
the matrix  $\bm D_{p_{id},q}$ we can compute the solution $\bm d_q$ of the inverse dynamics.
A more computationally efficient solution can be obtained as described in the following section.

\subsection{Inverse dynamics solved with forward substitution} \label{sec:matrixRNE}

In this section we prove that with suitable permutations we can compute $d_q$
which solves \eqref{eq:matRNEA3} with a computationally efficient algorithm,
which is the forward substitution as defined in \cite{golub2013}. The idea
is to build two find permutations $p_{id}$ and $q$ of the elements in 
$\mathcal C_{id}$ and $\mathcal D$ which lead to a lower triangular matrix
$D_{p_{id}, q}$ and then to solve \eqref{eq:matRNEA3} with a forward 
substitution. The permutations 
are inspired by the RNEA described in Section \ref{sec:RNE}. Let's first consider
the $q$ permutation of the elements in $\mathcal D$. We choose:
\begin{multline} \label{eq:qRNEA}
q  = \left[
f_1^{x}, \ddot {q}_1, \dots, f_{N_B}^{x}, \ddot {q}_{N_B},
a_1, \dots,  a_{N_B}, \right.\\
\left. 
f_{N_B}, {\tau}_{N_B}, \dots, f_1, {\tau}_1
\right].
\end{multline}
Let's also choose a permutation $p_{id}$ of the elements in $\mathcal C_{id}$ :
\begin{multline} \label{eq:pRNEA}
p_{id}  = \left[
c_{(\ref{eq:meas}f^x_1)}, c_{(\ref{eq:meas}\ddot q_1)} \dots
c_{(\ref{eq:meas}f^x_{N_B})}, c_{(\ref{eq:meas}\ddot q_{N_B})},
c_{(\ref{eq:NEdyn}a_1)} \dots  c_{(\ref{eq:NEdyn}a_{N_B})}, \right.\\
\left. 
c_{(\ref{eq:NEdyn}f_{N_B})}, c_{(\ref{eq:NEdyn}\tau_{N_B})},
\dots, 
c_{(\ref{eq:NEdyn}f_1)}, c_{(\ref{eq:NEdyn}\tau_1)}
\right].
\end{multline}

\begin{property} \label{prp:pRNEA}
The permuted matrix $\bm D_{p_{id},q}$ induced by the permutations defined in \eqref{eq:qRNEA}
and \eqref{eq:pRNEA} is lower triangular. 
\end{property}
 
\begin{proof}
Since $\bm D_{p_{id},q}$ is defined by blocks, we structure this proof 
by considering the blocks that constitute the matrix itself. First, we 
prove that the blocks in the main diagonal are diagonal matrices. These blocks
are:
\begin{align*}
 D_{c_\eqref{eq:ydfxi}, f^x_{i}} & = 1_6, &
 D_{c_\eqref{eq:yd2qi}, \ddot q_{i}} & = 1_{n_i}, &
 D_{c_\eqref{eq:ai}, a_{i}} & =-1_6, & \\
 D_{c_\eqref{eq:fi}, f_{i}} & =-1_6, &
 D_{c_\eqref{eq:taui}, \tau_{i}} & =-1_{n_i}, &  &
\end{align*}
which are indeed diagonal. We are left with proving that the blocks in the upper 
triangular part of $\bm D_{p_{id},q}$ are identically null. We consider the non-null
blocks in \eqref{eq:Dblocks} and prove that each block $D_{c,d}$ is positioned 
in the lower triangular part of $\bm D_{p_{id},q}$. This is equivalent to prove that 
if $D_{c,d} \neq 0$ then $i_c \geq i_d$, being $i_c$ and $i_d$ the position of $c$ 
and $d$ in the permutations $p_{id}$ and $q$, respectively. In the definitions
given in \eqref{eq:Dblocks}, we can neglect the 
diagonal blocks, which have been previously considered. We are left with:
\begin{align*}
  D_{c_\eqref{eq:ai}, a_{\lambda_i}}&, &
 i_{c_\eqref{eq:ai}} &= 2N_B+i, & 
 i_{a_{\lambda_i}} & = 2N_B+\lambda_i \\
 D_{c_\eqref{eq:ai}, \ddot {q_{i}}}&, &
 i_{c_\eqref{eq:ai}} &= 2N_B+i, & 
 i_{\ddot {q_{i}}} &= 2i \\
 D_{c_\eqref{eq:fi}, a_{i}}&, &
 i_{c_\eqref{eq:fi}} &= 5N_B-2i+1, & 
 i_{a_{i}} & = 2N_B+i \\
  D_{c_\eqref{eq:fi}, f^x_{i}}&, &
 i_{c_\eqref{eq:fi}} &= 5N_B-2i+1, & 
 i_{f^x_{i}} & = 2i-1 \\
  D_{c_\eqref{eq:fi}, f_{j}}&, &
 i_{c_\eqref{eq:fi}} &= 5N_B-2i+1, & 
 i_{f_j} & = 5N_B-2j+1 \\
  D_{c_\eqref{eq:taui}, f_{i}}&, &
 i_{c_\eqref{eq:taui}} &= 5N_B-2i+2, & 
 i_{f_i} & = 5N_B-2i+1.
\end{align*}
Easy computations can show that in these case $i_c \geq i_d$ 
considering a numbering scheme with $\lambda_i < i$ and $j > i$,
$\forall j \in \mu_i$ as described in \cite{Featherstone2008}.
\end{proof}

\subsection{Inverse dynamics solved with LU factorization} \label{sec:luRNE}

In Section \ref{sec:matrixRNE}, suitable permutations led to a lower triangular structure
for the matrix $\bm D$ in \eqref{eq:y_RNE}. After obtaining the 
lower triangular structure, the underlying linear system can
be solved with a forward substitution which is a computationally 
efficient algorithm to solve a linear system. In this section 
we are interested in computing these permutations
by solving the following problem.

\begin{problem} \label{prb:inversePerm}
Given arbitrary permutations $p$ and $q$ of the elements in $\mathcal C_{id}$
and $\mathcal D$, compute permutation matrices $P$ and $Q$
to obtain a triangular matrix $P \bm D_{p,q} Q$. 
\end{problem}

In consideration of Property \ref{prp:pRNEA}, Problem \ref{prb:inversePerm} has necessarily 
a solution. In other terms, $\bm D_{p,q}$ is triangularizable with permutations.

\begin{definition}[triangularizable with permutations] \label{def:triangularPerm}
A square matrix $A$ is triangularizable with permutations if there exists a permutation of
the rows and a permutation of the columns which result in a triangular matrix.
\end{definition} 

To solve Problem \ref{prb:inversePerm}, we resort to a classical problem in matrix analysis:
the sparse $LU$ factorization with minimum filling-in \cite[Section 11.1.9]{golub2013}. 

\begin{problem} \label{prb:minFillingLU}
Given a square matrix $A$ find the permutation matrices $P$ and $Q$, 
a lower triangular matrix $L$ and an upper triangular matrix 
$U$ such that $PAQ=LU$ and the number of filling-in 
(new nonzeros in $L$ and $U$ that are not present in $A$) 
is minimum. 
\end{problem}

\begin{remark} \label{rmk:triangularLU}
If $A$ is triangularizable with permutations, i.e. $A = \bar P \bar L \bar Q$ with $L$ triangular,
then the solution to problem \ref{prb:minFillingLU} is given by 
$P = \bar P^{-1}$, $Q = \bar Q^{-1}$, $L = \bar L$ and $U$ equal
to the identity matrix. In this case in fact, we can obtain zero filling-in
which by definition is the minimum number achievable. Under this considerations,
we can apply Problem \ref{prb:minFillingLU} to solve Problem \ref{prb:inversePerm}
\end{remark}

\begin{remark}
Deciding if a matrix is triangularizable according to definition 
\ref{def:triangularPerm} is $\mathcal{NP}$-complete \cite{fertin2015}.
Similarly, solving problem \ref{prb:minFillingLU} is also
$\mathcal{NP}$-complete \cite{Yannakakis1981}\footnote{Even though \cite{Yannakakis1981}
is cited several times \cite{Dongarra2001,Grigori2010} as a proof of the $\mathcal{NP}$-completeness 
of Problem \ref{prb:minFillingLU}, it has to be observed that it is not clear 
to the authors of the present paper how to extend the results in \cite{Yannakakis1981} to the 
case of non-symmetric positive definite matrices}. Available numerical tools for 
finding their solution are not guaranteed to reach the minimum. In solving
Problem \ref{prb:minFillingLU} we will use the unsymmetric-pattern 
multiFrontal method, as implemented in UMFPACK \cite{davis2004}.
\end{remark}

Going back to Problem \ref{prb:inversePerm}, we can take advantage 
of Property \ref{prp:pRNEA} to guarantee that $\bm D_{p,q}$ is always
triangularizable. Therefore, remark \ref{rmk:triangularLU} 
applied to $\bm D_{p,q}$ guarantees that solving Problem 
\ref{prb:minFillingLU} will give us also a solution to 
Problem \ref{prb:inversePerm}. Given the $\mathcal{NP}$-completeness
of the underlying problem, we are not guaranteed to find a solution
but numerical experiments\footnote{Experiments are avaialble here 
\url{https://github.com/iron76/bnt_time_varying/tree/master/experiments/computationalComplexity/RNEA}.}
conducted so far with UMFPACK shows that 
a solution is always found if $N_B \leq 100$. 
As to this concern, one-hundred can be considered a 
a practical upper-bound for robotic applications. 
 
\begin{remark}
Even though Problem \ref{prb:minFillingLU} is $\mathcal{NP}$-complete,
in practice the problem is solved 
once in a \emph{preliminary optimization} phase and its benefits can 
be exploited in the \emph{runtime computations}. The preliminary optimization
consists in solving Problem \ref{prb:minFillingLU} for a worst-case
sparsity pattern. The runtime computations instead consist in solving \eqref{eq:matRNEA3}
for different positions $q$ and velocities $\dot q$ exploiting the optimized permutations
computed previously. 

A solution of Problem \ref{prb:minFillingLU} can be used to compute 
permutations that allow to solve \eqref{eq:matRNEA3} efficiently. The
solution uses the sparsity pattern of $\bm D_{p,q}(q)$, i.e. 
the pattern of non-zero elements in the matrix. However, 
in practical applications we are interested in a sparsity pattern 
which somehow represents the sparsity of $\bm D_{p,q}(q)$ for
all possible values of $q$. As to this concern, we define 
the worst case sparsity pattern.

The basic observation is that the robot structure
(e.g. number of degrees of freedom, joint types, joint positions,
tree structure of the robot) does not change. As a consequence,
the underlying sparsity structure of \eqref{eq:matRNEA3}, i.e. 
the non-zero elements in $\bm D$, changes only for the state-dependent 
elements, i.e. those that depend on $q$ and $\dot q$. Looking at \eqref{eq:Dblocks},
the only state-dependent blocks are the transformations $\prescript{i}{}{X_{j}^*}$,
$\prescript{i}{}{X_{\lambda_i}}$ which depend on $q$. 
These sub-matrices do have a state-dependent sparsity structure but
for the purpose of this paper, we can consider the associated worst-case 
sparsity structure, i.e. if an element is non-zero for at least one 
value of $q$, then it is considered as a non-zero element in the associated
sparsity pattern. For classical joint types (revolute, prismatic, helical, 
cylindrical, planar, spherical and free-motion) the worst-case 
sparsity pattern can be easily computed by observing that the only 
$q$ dependent elements are either sines of cosines 
(see \cite[Table 4.1, page 79]{Featherstone2008}). 
These functions are zero only on
a countable number of configurations (and therefore on a subset
whose Lebesgue measure is zero) which are easy to enumerate.

\end{remark}
 
\section{Forward dynamics} \label{sec:fwdDyn}

The inverse dynamics problem consists in finding $\ddot {q}_1$, $\dots$, $\ddot {
q}_{N_B}$ which satisfies \eqref{eq:NEdyn}
given ${\tau_1}$, $\dots$, ${\tau_{N_B}}$,  $f_1^x$, $\dots$, $f_{N_B}^x$. 
In \cite{Featherstone2008} the problem is formulated
as the computation of the following function:
\begin{eqnarray}
\ddot {q} = \mbox{FwdD}(\textcolor{gray}{model, q, \dot {q}}, \tau, f_1^x, \dots, f_{N_B}^x).
\end{eqnarray}
Again, in the above equations we grayed out some variables that will not play a role in 
the following sections, and can be assumed either to be contestant ($model$) or measured ($q$, $\dot {q}$).

\subsection{Forward dynamics solved with the ABA} \label{sec:ABA}


The articulated-body algorithm \cite[ABA]{Featherstone2008} solves the forward dynamics 
problem in $O(N_B)$ computational complexity. The algorithm consists in the following steps.
First, the articulated body bias forces $p_i^A$ and the 
articulated body inertias $I_i^A$ are recursively computed iterating 
with $i$ = $N_B$, $\dots$, $1$ the following equations:

\footnotesize
\begin{align} 
\begin{split} \label{eq:ABAdyn_p}
    p_i^A ={}& \nu_i -  f_i^x + \sum_{j \in \mu_i} \prescript{i}{} {X_{j}^*} \Big\{ p_j^A 
+ I_j^a c_j   + \\            & +  I_j^A S_j \left(S_j^\top I_j^A S_j\right)^{-1} \left(\tau_j - 
S_j^\top p_j^A \right) \Big\},
\end{split} \tag{\theequation $p_i$}\\ \label{eq:ABAdyn_Ia}
I_j^a ={}& {I_j^A - I_j^A S_j 
\left(S_j^\top I_j^A S_j\right)^{-1}
S_j^\top I_j^A} \tag{\theequation $I_i^a$} \\ \label{eq:ABAdyn_IA}
I_i^A ={}& I_i + \sum_{j \in \mu_i} 
\prescript{i}{} {X_{j}^*} I_j^a \prescript{j}{} {X_{i}} \tag{\theequation $I_i^A$}
\end{align}
\normalsize

\noindent Then the following two equations are 
iterated with $i$ = $1$, $\dots$, $N_B$ and initial condition $a_0 = -a_g$.

\footnotesize
\begin{subequations}
\begin{align} \label{eq:ABAdyn_d2q}
\ddot {q}_i &=   \left( S^\top_i  I_i^A S_i \right)^{-1} 
\left\{ \tau_i - S^\top_i
\left[ I_{i}^A 
\left( \prescript{i}{}{X_{\lambda_i}} a_{\lambda_i} + c_{i} \right) + p_{i}^A \right] \right\} 
\tag{\theequation $\ddot q_i$} \\
\nonumber a_i    &= \prescript{i}{}{X_{\lambda_i}} a_{\lambda_i} + S_i \ddot {q}_i + c_i. 
\end{align}
\end{subequations}
\normalsize

\subsection{Forward dynamics solved with matrix inversion} \label{sec:matrixABA}

In solving the inverse dynamics, the fact that $\bm D_{p_{id},q}$ is lower triangular 
follows from the specific structure of the measurement equations \eqref{eq:meas}. Changing 
the measurement equations would compromise the lower triangularity of $\bm D_{p_{id},q}$. As 
a consequence, the associated linear system would not have the suitable structure to apply 
the forward substitution. Within this context, forward dynamics give a useful example. The 
measured variables for the forward dynamic case are $f^x_i$ and ${\tau}_i$. The 
latter in matrix notation can be expressed as follows:

\begin{subequations} \label{eq:meas2} 
\begin{align} 
\tau_i & = y_{\tau_i}, \label{eq:ytaui} \tag{\theequation $\tau_i$} 
\end{align}
\end{subequations}

These constraints, will extend the set of constraints $\mathcal C$. In particular, we
should define $\mathcal C_{fd} = \mathcal C \cup \{ c_\eqref{eq:ytaui}, c_\eqref{eq:ydfxi} \}_{i=1}^{N_B}$ and extend the definition of $D_{c,d}$ and $b_c$ as follows:
\begin{align*}
  D_{c_\eqref{eq:ytaui}, \tau_{i}} & = 1_{n_i}, & 
  b_{c_\eqref{eq:ytaui}} & = -y_{\tau_i},
\end{align*}
and $ D_{c,d} = 0$, $ b_{c} = 0$ otherwise. Given two permutations $p_{fd}$ and
$q$ of the elements in $\mathcal C_{fd}$ and $\mathcal D$, a solution of 
the forward dynamics can be computed as the unique solution $\bm d_{q}$ of the following
liner system: 
\begin{equation} \label{eq:matABA}
 \bm D_{p_{fd},q}(q) \bm d_{q} + \bm b_{p_{fd}} (q,{\dot q}) = 0, 
\end{equation}

\subsection{Forward dynamics solved with forward substitution} \label{sec:optimizedMatrixABA}

Similarly to what observed in the inverse dynamic case, we might try to find permutations 
for $p_{fd}$ of the elements in $\mathcal C_{fd}$ and $q$ of the elements in $\mathcal D$ 
to obtain a matrix $\bm D_{p_{fd}, q}$ somehow simple to invert. Specifically, we 
might think that the articulated-body algorithm \cite[the ABA]{Featherstone2008} 
presented in Section \eqref{sec:ABA},
could be translated into suitable permutations that transform $\bm D_{p_{fd}, q}$
into a lower triangular matrix. Unfortunately, the ABA algorithm is instead something more than 
a permutation as discussed in the following. 

\begin{property} \label{prp:ABAtrinagularisation}
There exist: 
\begin{itemize}
 \item[-] a permutation $p_{fd}$ of the elements in $\mathcal C_{fd}$; 
 \item[-] a permutation $q$ of the elements in $\mathcal D$;
 \item[-] matrices $W^R_{q_1, q_2}$ defined for $q_1, q_2 \in \mathcal D$;
 \item[-] matrices $W^L_{p_1,p_2}$ defined for $p_1, p_2 \in \mathcal C_{fd}$;
\end{itemize}
such that:
$$\bm W^L_{p_{fd},p_{fd}} \bm D_{p_{fd},q} \bm W^R_{q,q}$$
is lower triangular. These quantities lead to the following matrix reformulation of
the ABA algorithm:
$$\bm W^L_{p_{fd},p_{fd}} \bm D_{p_{fd},q} \bm W^R_{q,q} 
 \bm d_{q} + \bm b_{p_{fd}} (q,{\dot q}) = 0.
$$
\end{property}

The proof of the preposition above is constructive and can be found in
Appendix \ref{app:ABAtrinagularisation}.

\subsection{Forward dynamics solved with LU factorization} \label{sec:luAB}

\begin{figure} 
\centering 
\includegraphics[height=5.5cm]{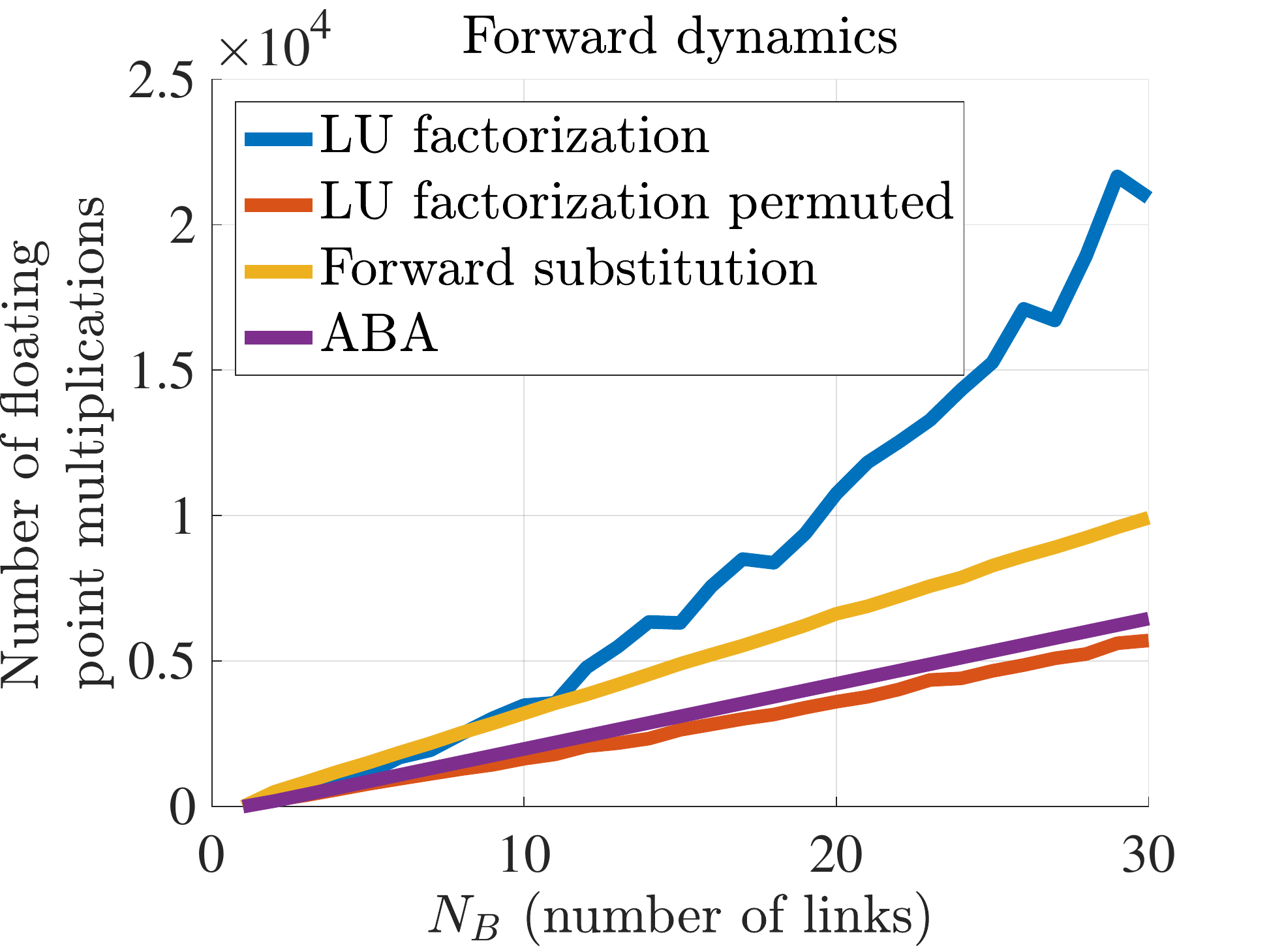} 
\caption{Comparison of the different proposed algorithms for 
solving the forward dynamics of a serial chain with $N_B$ links.}
\label{fig:luAB}
\end{figure}

Again, part of the computational optimizations of the ABA
proposed in Section \ref{sec:ABA} and revisited in Section \ref{sec:optimizedMatrixABA}
consists in row and column permutations. In this section we 
will suggest an algorithm for simplifying the forward dynamics 
computations by leveraging the intrinsic sparsity of the 
underlying matrices \eqref{eq:matABA} and pre-computing suitable 
row and column permutations that will reduce the underlying 
computational cost. The idea consists in computing the worse 
case sparsity pattern for $\bm D_{p_{fd},q}(q)$ in \eqref{eq:matABA}.
Solving Problem \ref{prb:minFillingLU} with this sparsity pattern
will give the row-column permutations $P$ and $Q$ suitable for solving 
\eqref{eq:matABA}. At run-time regardless of the specific $q$
the idea is to perform the sparse $LU$ factorization on 
$P \bm D_{p_{fd},q}(q) Q$ followed by a backward substitution on $U$
and a forward substitution on $L$. Figure \ref{fig:luAB}
shows the computational cost of this solution against 
the algorithms proposed in Section \ref{sec:ABA} (ABA)
and Section \ref{sec:optimizedMatrixABA} (matrix reformulation
of the ABA). The computational cost of ABA is the one reported 
in \cite[Page 202]{Featherstone2008} and corresponds to the 
algorithmic solution proposed in \cite{Brandl1986}. The
other computational costs are numerically computed with 
software which is available open source\footnote{The
software for these computations are available here:
\url{https://github.com/iron76/bnt_time_varying/tree/master/experiments/computationalComplexity/ABA/serial/compare}.}.

\section{Dynamic equations and LU-factorization} \label{sec:LUgeneric}

In the previous sections, we have seen how the RNEA and 
the ABA are computationally efficient solutions of the 
inverse and forward dynamics problems. The reduction
of the computational costs is obtained by suitable
permutations of the matrices $\{D_{c,d}\}_{c \in \mathcal
C_{id}, d \in \mathcal D}$ and $\{D_{c,d}\}_{c \in \mathcal
C_{fd}, d \in \mathcal D}$. In this section we consider the
problem of finding similar permutations for a wider class
of problems. Inspired by the matrix representations 
\eqref{eq:matRNEA3} and \eqref{eq:matABA} of the inverse and 
forward dynamics respectively, we consider a generic estimation
problem consisting in computing the solution $\bm d$ of
the following linear system:

\begin{equation} \label{eq:genericDynamic}
 \underbrace{\begin{bmatrix}
  \bm D_{p,q}(q) \\
  \bm D_Y(q)
 \end{bmatrix}}_{\triangleq \bm D(q)}
  \bm d + 
   \underbrace{\begin{bmatrix}
    \bm b_{p} (q,{\dot q})\\
    \bm b_Y(q, \dot q)
 \end{bmatrix}}_{\triangleq \bm b(q, \dot q)} = 0, 
\end{equation}
obtained by combining \eqref{eq:matRNEA2} with a generic 
measurement equation $\bm D_Y(q) \bm d + \bm b_Y(q, \dot q) = 0$.
In Section \ref{sec:matrixRNE} and Section \ref{sec:optimizedMatrixABA}
respectively, suitable permutations led to a lower triangular structure
for the matrix $\bm D$. The underlying linear system could then
be solved with a forward substitution. However, the proposed 
solutions rely on the specific structure of the problem and 
therefore extending them to solve \eqref{eq:genericDynamic}
is non trivial. The algorithm proposed in Section \ref{sec:luRNE} and 
Section \ref{sec:luAB} seems more suitable to simplify
the computational complexity of solving \eqref{eq:genericDynamic}  
in its generic form (i.e. beyond the inverse and forward dynamics cases). 

As an example, we consider a relevant case study: the inverse
dynamic computation of a humanoid robot standing on two feet.
The specificity of this problem is that the robot has two external 
wrenches applied at the feet. Differently from the inverse dynamic
computations presented in Section \ref{sec:RNE}, we assume that these
external wrenches are unknown. Using the notation used in previous sections,
these external wrenches can be denoted $f_1^x$ and $f_{N_B}^x$ by labelling 
with $1$ and $N_B$ the right and the left foot respectively. We therefore 
consider the following problem:

\begin{eqnarray} \label{eq:invDynNao}
\tau = \mbox{InvD}(\textcolor{gray}{model, q, \dot {q},} \ddot {q}, f_2^x, \dots, f_{N_B-1}^x),
\end{eqnarray}

\noindent
where we explicitly indicated that $f_1^x$ and $f_{N_B}^x$ are unknown 
even if they act on the system. The considered mechanical system is 
a free-floating articulated
rigid body subject to constraints. 
Unfortunately solving \eqref{eq:invDynNao} is ill-posed, i.e. 
given $q, \dot {q}, \ddot {q}, f_2^x, \dots, f_{N_B-1}^x$ 
there exist multiple $\tau$, $f_1^x$, $f_{N_B}^x$ satisfying \eqref{eq:NEdyn}.
To estimate $\tau$ we could try using additional 
measurements. We hereafter consider quite a common sensor distribution, 
nominally four load cells located on each foot\footnote{We first prove that
these sensors are not yet sufficient to solve the inverse dynamics. 
Then we propose a different set of sensors which results in a 
well posed inverse dynamics problem.}. 
These sensors are available in the NAO robot \cite{Gouaillier2009}, 
in the QRIO robot \cite{Ishida2004} and the Atlas \cite{Kuindersma2016}.
These sensors correspond to a three-axes force-torque sensor,
i.e. they measure the net force orthogonal to 
the plane of the each foot and the projection of the torque on the same
plane. As it is often the case, if we choose the left and right feet reference frames with the 
$x$ and $y$ axis aligned to the foot plane we have: 

\footnotesize
\begin{subequations} \label{eq:decomp}
 \begin{align}
f_1^x &= Y y_{lf}
+
H h_{lf}, &
f_{N_B}^x & = Y y_{rf}
+
H h_{rf}, \\ \label{eq:defH}
Y & \triangleq \begin{bmatrix}
0 & 0 & 0 \\
0 & 0 & 0 \\
0 & 0 & 1 \\
1 & 0 & 0 \\
0 & 1 & 0 \\
0 & 0 & 0 
\end{bmatrix}, &
H & \triangleq \begin{bmatrix}
1 & 0 & 0 \\
0 & 1 & 0 \\
0 & 0 & 0 \\
0 & 0 & 0 \\
0 & 0 & 0 \\
0 & 0 & 1 
\end{bmatrix},
\end{align}
\end{subequations}
\normalsize

\noindent
with $y_{lf},y_{rf} \in \mathbb R^3$ being the measured three-axes force-torque
and $h_{lf},h_{rf} \in \mathbb R^3$ being the non-directly measured components 
of the contact wrench $f_1^x$. We are now considering the following problem:

\begin{eqnarray} \label{eq:invDynNao2}
\tau = \mbox{InvD}(\textcolor{gray}{model, q, \dot {q},} \ddot {q}, f_2^x, \dots, f_{N_B-1}^x, y_{lf}, y_{rf}).
\end{eqnarray}

\noindent
To understand if this problem is well-posed, it is convenient to resort to a specific 
form of equation \eqref{eq:NEdyn}. This formulation is detailed in \cite[eq. (41)]{Traversaro2017}
and it combines the floating-base dynamics with the joint dynamics.
We consider here a simplified version obtained by choosing the
left foot as the base frame and choosing $q$ a local parametrization of the robot pose (composed by the 
free-floating six-dimensional configuration and the joint 
$n$-dimensional configuration). We have:

\begin{equation} \label{eq:fb}
M(q) \ddot q + h(\dot q, q)  = \begin{bmatrix}
\tau \\
0
\end{bmatrix} + \begin{bmatrix}
\prescript{0}{}{J^\top_{1}}  \\ 1_6
 \end{bmatrix}
 f_1^x + \sum_{j = 2}^{N_B}
\begin{bmatrix}
\prescript{0}{}{J^\top_{j}} \\   \prescript{0}{}{X_{j}^*}
 \end{bmatrix} f_j^x .
\end{equation}

\noindent
Using \eqref{eq:decomp} in \eqref{eq:fb} and grouping the unobservable/observable
quantities as in problem \eqref{eq:invDynNao2}, we obtain:

\begin{multline}
Y(q, \dot {q}, \ddot {q}, f_2^x, \dots, f_{N_B-1}^x, y_{lf}, y_{rf}) = \\ = \begin{bmatrix}
\tau  \\
0
\end{bmatrix} + \begin{bmatrix}
\prescript{0}{}{J^\top_{1}} H h_{lf}  \\ H h_{lf}
 \end{bmatrix}
 + 
\begin{bmatrix}
\prescript{0}{}{J^\top_{N_B}} H h_{rf} \\  \prescript{0}{}{X_{N_B}^*} H h_{rf}
 \end{bmatrix}.
\end{multline}

Understanding if the inverse dynamic problem with these measurements is
well-posed boils down to understanding if the following matrix is invertible:

\begin{eqnarray}
\begin{bmatrix}
1_n & \prescript{0}{}{J^\top_{1}} H & \prescript{0}{}{J^\top_{N_B}} H  \\
0_{6 \times 6} &  H  & \prescript{0}{}{X_{N_B}^*} H 
\end{bmatrix}.
\end{eqnarray}

\noindent
Given the upper triangular form of this matrix, we are left with proving 
the invertibility of $\left[ \begin{smallmatrix} H , & 
\prescript{0}{}{X_{N_B}^*} H \end{smallmatrix} \right]$. We are interested
in guaranteeing the invertibility of this matrix for any $\prescript{0}{}{X_{N_B}^*}$,
i.e. regardless of the feet relative pose. 

\begin{property}\label{prp:singularityH}
Let $H = \left[ \begin{smallmatrix} H_f^\top & H^\top_\mu \end{smallmatrix} \right]^\top$
with $H_f \in \mathbb{R}^{3 \times 3}$, $H_\mu \in \mathbb{R}^{3 \times 3}$. If
$H_f$ is singular, then there always exists $\prescript{0}{}{X_{N_B}^*}$ which
makes $\left[ \begin{smallmatrix} H , & 
\prescript{0}{}{X_{N_B}^*} H \end{smallmatrix} \right]$ singular.
\end{property}
 
\begin{proof} 
Using the structure of $\prescript{0}{}{X_{N_B}^*}$ and $H$ we obtain:
\begin{eqnarray*}
\begin{bmatrix}
H & \prescript{0}{}{X_{N_B}^*} H 
\end{bmatrix}
& = & 
\begin{bmatrix}
 H_f & R H_f\\
 H_\mu & p \times R H_f + RH_\mu 
 \end{bmatrix} \\
& \stackrel{R = 1_3}{=} & 
\begin{bmatrix}
 H_f & 0_{3 \times 3}\\
 H_\mu & p \times H_f  
\end{bmatrix} 
\begin{bmatrix}
 1_3 & 1_3 \\
 0_{3 \times 3} & 1_3
 \end{bmatrix},
\end{eqnarray*}
and the results follows by observing that the multiplicand matrix on the left
is singular (the first three lines are linearly dependent) and by using the
fact that the rank of a product is always less or equal the rank of the
multiplied matrices. \end{proof} 

\begin{remark}
Using Property \ref{prp:singularityH} with definitions \eqref{eq:defH}
we can conclude that \eqref{eq:invDynNao2} is an ill-posed problem. In
other terms, the inverse dynamics problem cannot be solved in the case
of a humanoid robot standing on the two feet with four load cells on 
each foot.
\end{remark}

\begin{figure} 
 \centering
\includegraphics[height=5.5cm]{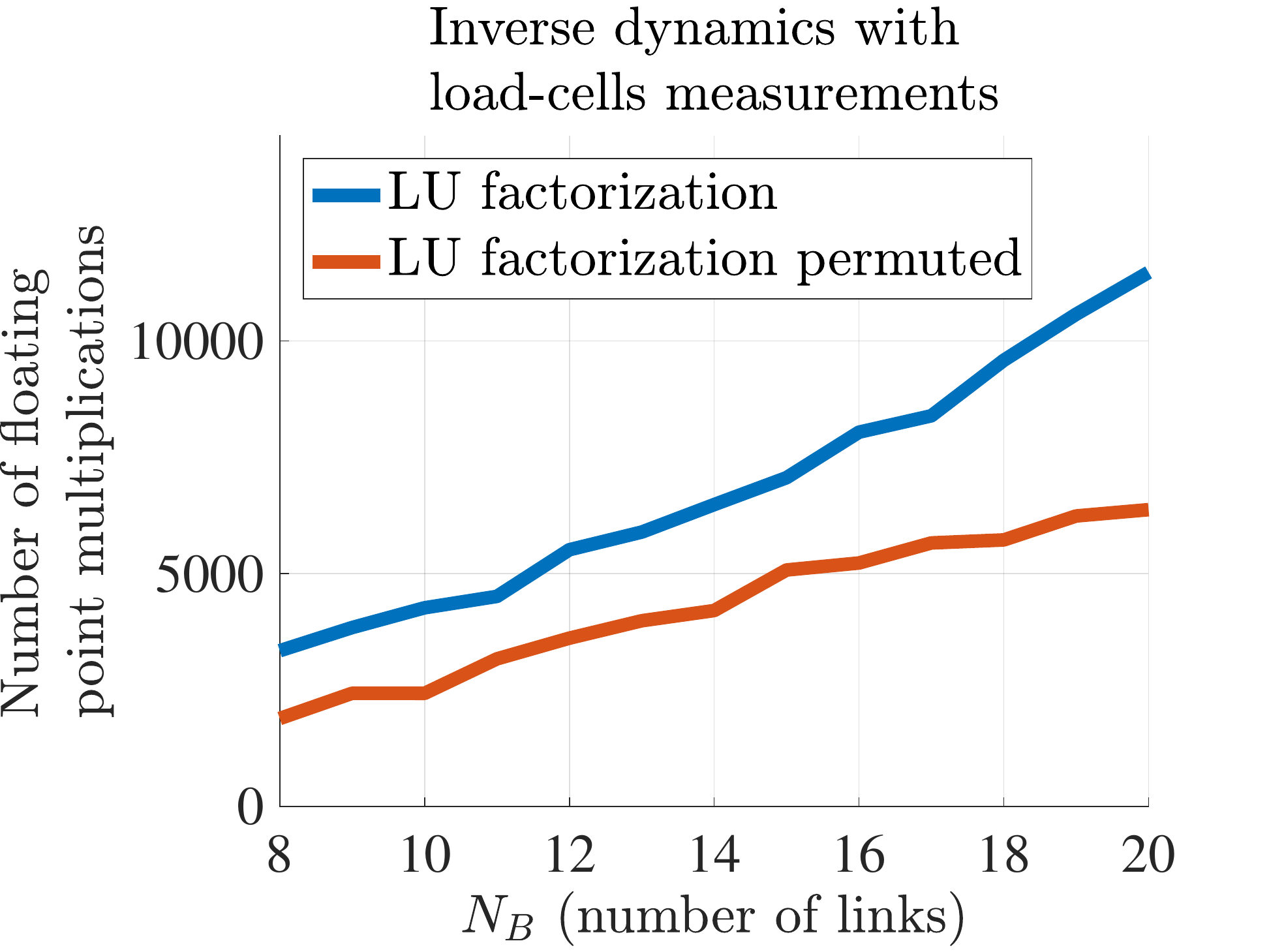} 
\caption{Comparison of the different proposed algorithms for 
solving the inverse dynamics (i.e. joint torque estimation)
of a serial chain with $N_B$ links. Considered measurements
are load-cells at extremal links.}
\label{fig:luSIE}
\end{figure}

Additional assumptions are needed to compute the inverse dynamics. 
A first realistic assumption in certain applications is to assume 
that the robot is standing in a very slippery surface and therefore
the tangential forces along the $x$ and $y$ axis are negligible. 
In this case:

\begin{subequations} \label{eq:decomp2}
 \begin{align}
f_1^x &= Y y_{lf}
+
H h_{lf}, &
f_{N_B}^x & = Y y_{rf}
+
H h_{rf}, \\ \label{eq:defH2}
Y & \triangleq \begin{bmatrix}
1 & 0 & 0 & 0 & 0  \\
0 & 1 & 0 & 0 & 0  \\
0 & 0 & 1 & 0 & 0  \\
0 & 0 & 0 & 1 & 0  \\
0 & 0 & 0 & 0 & 1  \\
0 & 0 & 0 & 0 & 0  
\end{bmatrix}, &
H & \triangleq \begin{bmatrix}
0 \\
0 \\
0 \\
0 \\
0 \\
1 
\end{bmatrix},
\end{align}
\end{subequations}

\noindent
with $y_{lf},y_{rf} \in \mathbb R^3$ being the measured contact forces and the 
measured torques on the $x$-$y$ plane and $h_{lf},h_{rf} \in \mathbb R$ being
the torques on the $z$ axis. In this case it was numerically 
observed\footnote{\url{https://github.com/iron76/bnt_time_varying/tree/master/experiments/computationalComplexity/SIE}}
that the associated inverse dynamics problem is solvable. Fig.~\ref{fig:luSIE}
shows a comparison of the number of floating point operations necessary
to solve this specific inverse dynamics problem with or without
a sparse $LU$ factorization. Remarkably this case cannot be solved
with classical algorithms (e.g. hybrid dynamics \cite{Featherstone2008}).

\appendix

\subsection{The articulated body equation of motion} \label{sec:articulatedBodyEquation}

The idea consists in recursively computing 
($i$ = $N_B$, $\dots$, $1$) the quantities $p_i^A$ (articulated body bias forces)
and $I_i^A$ (articulated body inertias) which satisfy 
the articulated body equation of motion: 

\begin{align} \label{eq:AB}
 f_i = I_i^A a_i + p_i^A. \tag{\theequation $f_i^{ab}$}
\end{align}

\noindent
For defining these quantities, let's start by considering $i$ such that 
$\mu_i = \emptyset$. In this case \eqref{eq:fi}
gives $f_i =  I_i a_i + \nu_i  -  f_i^x$
and therefore $I_i^A = I_i$ and $p_i^A = \nu_i  -  f_i^x$.
Recursively, let's assume that $I_j^A$ and $p_j^A$ have
been defined for every $j \in \mu_i$ (this time non-empty)
and let's find suitable expressions for $I_i^A$ and $p_i^A$. 
This is achieved by a five step procedure.

\noindent
\emph{Step 1 - $S_1$}. Replace \eqref{eq:AB} in \eqref{eq:NEdyn}:

\begin{subequations} \label{eq:ABdyn} 
\begin{align} \label{eq:ABA_ai}
a_i    &= \prescript{i}{}{X_{\lambda_i}} a_{\lambda_i} + S_i \ddot {q}_i + c_i,  \tag{\theequation $a_i$} \\
\label{eq:AB_taui}
 \tau_i  & = S^\top_i \left( I_i^A a_i + p_i^A\right) \tag{\theequation $\tau_i$}\\
 \label{eq:AB_with_aj}
 f_i & = I_i a_i + \nu_i -  f_i^x  + \sum_{j \in \mu_i}
  \prescript{i}{}{X_{j}^*} \left( I_j^A a_j + p_j^A \right) \tag{\theequation $f_i$}.
\end{align}
\end{subequations}
 

\noindent
The latter is almost the equation we need for the recursive definition of $I_i^A$ and $p_i^A$
if only we could write $a_j$ as a function of $a_i$. This is achieved with the following steps.

\noindent
\emph{Step 2 - $S_2$}. Substitute \eqref{eq:ABA_ai} in \eqref{eq:AB_taui}:

\begin{align} \label{eq:ABAdyn_taui}
\tau_j  = S^\top_j \left[ I_j^A 
\left( \prescript{j}{}{X_{\lambda_j}} a_{\lambda_j} + S_j \ddot {q}_j + c_j \right)
+ p_j^A\right].
\end{align}

\noindent
\emph{Step 3 - $S_3$}. Multiply the previous equation by the inverse of $S^\top_j  I_j^A S_j$
to obtain \eqref{eq:ABAdyn_d2q} for $\ddot {q}_j$: 

\footnotesize
\begin{eqnarray*} 
\ddot {q}_j & = & \left( S^\top_j  I_j^A S_j \right)^{-1} 
\left\{ \tau_j - S^\top_j
\left[ I_j^A 
\left( \prescript{j}{}{X_{\lambda_j}} a_{\lambda_j} + c_j \right) + p_j^A \right] \right\}\\
& \stackrel{j \in \mu_i}{=} & \left( S^\top_j  I_j^A S_j \right)^{-1} 
\left\{ \tau_j - S^\top_j
\left[ I_j^A 
\left( \prescript{j}{}{X_{i}} a_{i} + c_j \right) + p_j^A \right] \right\}, 
\end{eqnarray*}
\normalsize

\noindent 
\emph{Step 4 - $S_4$}. Substitute the last equation in \eqref{eq:ABA_ai}:

 \begin{multline*} 
a_i = \prescript{i}{}{X_{\lambda_i}} a_{\lambda_i} + c_i + S_i 
\left( S^\top_i  I_i^A S_i \right)^{-1} \\
\left\{ \tau_i - S^\top_i
\left[ I_i^A 
\left( \prescript{i}{}{X_{\lambda_i}} a_{\lambda_i} + c_i \right) + p_i^A \right] \right\}.
 \end{multline*}

\noindent 
which evaluated with the substitution $i \rightarrow j \in \mu_i$ leads to:

\begin{multline} \label{eq:AB_ai}
a_j = \prescript{j}{}{X_{i}} a_{i} + c_j + S_j 
\left( S^\top_j  I_j^A S_j \right)^{-1} \\
\left\{ \tau_j - S^\top_j
\left[ I_j^A 
\left( \prescript{j}{}{X_{i}} a_{i} + c_j \right) + p_j^A \right] \right\}.
 \end{multline}

\noindent
\emph{Step 5 - $S_5$}. Substitute $a_j$ in \eqref{eq:AB_with_aj} with its expression in \eqref{eq:AB_ai}:

 \begin{multline} \label{eq:AB_fi}
  f_i = \left( I_i + \prescript{i}{}{X_{j}^*} I_j^A  \prescript{j}{}{X_{i}} \right)
  a_i + \nu_i -  f_i^x  + \\
+ \sum_{j \in \mu_i} \prescript{i}{}{X_{j}^*} I_j^A S_j \left( S^\top_j  I_j^A S_j \right)^{-1} \tau_j \\
+ \prescript{i}{}{X_{j}^*} \left[ {1 - I_j^A S_j 
\left(S_j^\top I_j^A S_j\right)^{-1}
S_j^\top} \right] p_j^A \\
 - \left[ \prescript{i}{}{X_{j}^*} I_j^A S_j \left( S^\top_j  I_j^A S_j \right)^{-1} S^\top_j I_j^A \prescript{j}{}{X_{i}}\right] a_{i} \\
 + \prescript{i}{}{X_{j}^*} \left[ {I_j^A - I_j^A S_j 
\left(S_j^\top I_j^A S_j\right)^{-1}
S_j^\top I_j^A} \right]  c_j.
 \end{multline}

\noindent
and enforcing $f_i = I_i^A a_i + p_i^A$ leads to the definitions in \eqref{eq:ABAdyn_p},
\eqref{eq:ABAdyn_Ia} and \eqref{eq:ABAdyn_IA}.

\subsection{Proof of Property \ref{prp:ABAtrinagularisation}} \label{app:ABAtrinagularisation}

We hereafter assume that the articulated body inertias $I_i^A$ ($i$ = $1$, $\dots$, $N_B$) 
have been computed. The idea is to follow the steps presented in Section \ref{sec:articulatedBodyEquation}
to compute the sub-blocks of the matrices $\bm W^L$ and $\bm W^R$. The latter is defined 
as a matrix equivalent of \emph{Step 1}. The former as $\bm W^L = W^{L,4} W^{L,3} W^{L,2}
W^{L,1}$ with $W^{L,1}$ representing \emph{Step 2}, $W^{L,2}$ representing \emph{Step 3},
$W^{L,3}$ representing \emph{Step 4} and $W^{L,4}$ representing \emph{Step 5}.

\noindent
\emph{Step 1 - $S_1$}. We compute a matrix $W^R$ which multiplied 
by $d_q$ replaces $f_i$ with $p_i^A = f_i - I_i^A a_i$.
As usual we define $W^R$ by its blocks $W^R_{q_1, q_2}$ 
with $q_1, q_2 \in \mathcal D$. We have:

\begin{align}
 W^R_{f_{i}, a_{i}} & = I_i^A, &
 W^R_{q, q} & = 1 \quad \forall q \in \mathcal D, &
\end{align}

\noindent 
and $W^R_{q_1, q_2} = 0$ otherwise. 
\begin{remark}
Given two permutations $p$ and $q$ of the elements in $\mathcal C$ and $\mathcal D$,
if $D_{p, q}$ represents \eqref{eq:NEdyn}, $D_{p, q} W^R_{q, q}$
represents \eqref{eq:ABdyn}. Therefore, in the following
$\{ c_\eqref{eq:ai}$, $c_\eqref{eq:taui}$, $c_\eqref{eq:fi},
 \}_{i=1}^{N_B}$ can be read as $\{ c_\eqref{eq:ABA_ai}$, $c_\eqref{eq:AB_taui}$,
 $c_\eqref{eq:AB_with_aj} \}_{i=1}^{N_B}$.
\end{remark}
 
\noindent
\emph{Step 2 - $S_2$}. We define 
$W^{L,1}$ which left multiplies $D_{p, q} W^R_{q, q}$ to 
substitute \eqref{eq:ABA_ai} in \eqref{eq:AB_taui}. The exceptions
to $W^{L,1}_{p_1, p_2} = 0$ are:

\begin{align}
 W^{L,1}_{c_\eqref{eq:taui}, c_\eqref{eq:ai}} & = S_i^\top I_i^A, &
 W^{L,1}_{p, p} & = 1 \quad \forall p \in \mathcal C_{fd}, &
\end{align}

\noindent
\emph{Step 3 - $S_3$}. We define 
$W^{L,2}$ which left multiplies $W^{L,1}_{p,p}D_{p, q} W^R_{q, q}$ to 
multiply \eqref{eq:ABAdyn_taui} by the inverse of $S^\top_j I_j^A S_j$.
The exceptions to $W^{L,2}_{p_1, p_2} = 0$ are:

\begin{align}
 W^{L,2}_{c_\eqref{eq:taui}, c_\eqref{eq:taui}} & = \left(S^\top_j I_j^A S_j\right)^{-1}, &
 W^{L,2}_{p, p} & = 1 \mbox{ } \forall p \in \mathcal C_{fd}, q \neq \tau_{i}&
\end{align}

\noindent
\emph{Step 4 - $S_4$}. We define 
$W^{L,3}$ which left multiplies $W^{L,2}_{p,p} W^{L,1}_{p,p}D_{p, q} W^R_{q, q}$ to 
replace the occurrences of $\ddot {q}_j$ in \eqref{eq:ABA_ai} with their expression obtained 
in the previous step. The exceptions to $W^{L,3}_{p_1, p_2} = 0$ are:

\begin{align}
 W^{L,3}_{c_\eqref{eq:ai}, c_\eqref{eq:taui}} & = -S_i, &
 W^{L,3}_{p, p} & = 1 \quad \forall p \in \mathcal C_{fd}. &
\end{align}

\noindent
\emph{Step 5 - $S_5$}. We define 
$W^{L,4}$ which left multiplies $W^{L,3}_{p,p} W^{L,2}_{p,p} 
W^{L,1}_{p,p}D_{p, q} W^R_{q, q}$ to 
replace the occurrences of $a_j$ in \eqref{eq:AB_with_aj} with their expression obtained 
in the previous step. The exceptions to $W^{L,4}_{p_1, p_2} = 0$ are:

\begin{align}
 W^{L,4}_{c_\eqref{eq:fi}, c_{(\ref{eq:NEdyn}a_j)}} & = \prescript{i}{}{X_{j}^*} I_j^A, &
 W^{L,4}_{p, p} & = 1 \quad \forall p \in \mathcal C_{fd}. &
\end{align}

We are left with the definition of the permutations. Let's first consider
the $q$ permutation of the elements in $\mathcal D$. We choose:
\begin{multline}
q  = \left[
f_1^{x}, {\tau}_1, \dots, f_{N_B}^{x}, {\tau}_{N_B},
f_{N_B}, \dots, f_1, \right.\\
\left. 
a_1, \dots,  a_{N_B}, \ddot {q}_1, \dots, \ddot {q}_{N_B}
\right].
\end{multline}
Let's also choose a permutation $p_{fd}$ of the elements in $\mathcal C_{fd}$ :
\begin{multline}
p_{fd}  = \left[
c_{(\ref{eq:meas}f^x_1)}, c_{(\ref{eq:NEdyn}\tau_{N_B})} \dots
c_{(\ref{eq:meas}f^x_{N_B})}, c_{(\ref{eq:NEdyn}\tau_1)},
c_{(\ref{eq:NEdyn}f_{N_B})} \dots  c_{(\ref{eq:NEdyn}f_1)}, \right.\\
\left. 
c_{(\ref{eq:NEdyn}a_1)},
\dots, 
c_{(\ref{eq:NEdyn}a_{N_B})},
c_{(\ref{eq:meas2}\ddot q_1)},
\dots, 
c_{(\ref{eq:meas2}\ddot q_{N_B})}
\right].
\end{multline}